\documentclass{article}

\usepackage{float}
\usepackage{microtype}
\usepackage{graphicx}
\usepackage{subcaption}
\usepackage{diagbox}
\usepackage{booktabs}
\usepackage{multirow}
\usepackage{tabularx}
\usepackage{csquotes}

\usepackage{hyperref}

\usepackage[preprint]{icml2026}
\usepackage{amsmath}
\usepackage{amssymb}
\usepackage{mathtools}
\usepackage{amsthm}
\usepackage{nicefrac}

\usepackage[capitalize,noabbrev]{cleveref}

\theoremstyle{plain}

\theoremstyle{definition}

\theoremstyle{remark}

\usepackage[textsize=tiny]{todonotes}

\icmltitlerunning{Probing GNN Activation Patterns Through Graph Topology}

\begin{document}

\twocolumn[
  \icmltitle{Probing Graph Neural Network Activation Patterns Through Graph Topology}

  % You can specify symbols, otherwise they are numbered in order. Ideally, you
  % should not use this facility. Affiliations will be numbered in order of
  % appearance and this is the preferred way.
  \icmlsetsymbol{equal}{*}

  \begin{icmlauthorlist}
    \icmlauthor{Floriano Tori}{equal,vub}
    \icmlauthor{Lorenzo Bini}{equal,ugen}
    \icmlauthor{Marco Sorbi}{ugen_v2}
    \icmlauthor{Stéphane Marchand-Maillet}{ugen}
    \icmlauthor{Vincent Ginis}{vub,harvard}
  \end{icmlauthorlist}

  \icmlaffiliation{ugen}{Department of Computer Science, University of Geneva, Geneva, Switzerland}
  \icmlaffiliation{vub}{Business Technology and Operations, Vrije Universiteit Brussels, Brussels, Belgium}
  \icmlaffiliation{ugen_v2}{Research Institute for Statistics and Information Science, Geneva, Switzerland}

  \icmlaffiliation{harvard}{School of Engineering and Applied Sciences, Harvard University}
  \icmlcorrespondingauthor{Floriano Tori}{Floriano.tori@vub.be}
  \icmlcorrespondingauthor{Lorenzo Bini}{Lorenzo.bini@unige.ch}

  % You may provide any keywords that you find helpful for describing your
  % paper; these are used to populate the "keywords" metadata in the PDF but
  % will not be shown in the document
  \icmlkeywords{Machine Learning, Graphs Neural Networks, Activations, Curvature}

  \vskip 0.3in
]

\printAffiliationsAndNotice{\icmlEqualContribution}
\begin{abstract}
Curvature notions on graphs provide a theoretical description of graph topology, highlighting bottlenecks and denser connected regions. Artifacts of the message passing paradigm in Graph Neural Networks, such as oversmoothing and oversquashing, have been attributed to these regions. However, it remains unclear how the topology of a graph interacts with the learned preferences of GNNs. Through Massive Activations, which correspond to extreme edge activation values in Graph Transformers, we probe this correspondence. Our findings on synthetic graphs and molecular benchmarks reveal that MAs do not preferentially concentrate on curvature extremes, despite their theoretical link to information flow. On the Long Range Graph Benchmark, we identify a systemic \textit{curvature shift}: global attention mechanisms exacerbate topological bottlenecks, drastically increasing the prevalence of negative curvature. Our work reframes curvature as a diagnostic probe for understanding when and why graph learning fails.
\end{abstract}

\section{Introduction}
The graph structure of data, described by the set of nodes and edges, plays a crucial role in Graph Neural Networks (GNNs). Most standard GNNs are based on the message-passing paradigm \cite{scarselli2008graph, kipf2016semi}, where nodes update their representations by aggregating information from their local neighbors. Although effective for a range of tasks, this local aggregation mechanism faces fundamental limitations in capturing long-range dependencies. In addition, this paradigm is subject to phenomena such as oversquashing \cite{alon2020bottleneck,black2023understanding,di2023does} and oversmoothing \cite{li2018deeper, oono2019graph}.

Oversquashing, described as \textit{“exponentially growing information in (a) fixed-size vector”} \cite{alon2020bottleneck}, is a phenomenon related to the computational tree that originates from message-passing. This phenomenon was later theoretically linked to the topological bottleneck of the underlying graph structure, where negatively curved edges, indicative of bottleneck regions, supposedly hindered information flow \cite{toppingunderstanding}. To mitigate this, works therefore proposed curvature-based rewiring strategies \cite{toppingunderstanding, karhadkar2022fosr} aimed at flattening the geometry of the graph to alleviate these geometric bottlenecks. However, the relationship between topological and computational bottlenecks has recently been under scrutiny \cite{torieffectiveness,arnaiz2025oversmoothing}.

In parallel, Graph Transformers (GTs) \cite{dwivedi2020generalization, kreuzer2021rethinking} have emerged as a powerful alternative, promising to solve the long-range dependency problem through global attention mechanisms. By allowing direct communication between any pair of nodes, GTs theoretically bypass topological constraints entirely, effectively operating on a fully connected latent graph. However, the interplay between these two solutions, topological rewiring, and global attention, remains poorly understood. Does the learned attention mechanism in practice identify these topological bottlenecks? 

In this work, we investigate this question by analyzing the \enquote{Massive Activations} (MAs) \cite{bini2024massive} of Graph Transformers through the lens of \textit{Balanced Forman curvature} (BFc) \cite{toppingunderstanding}. Massive Activations, defined as extreme edge activation values, serve as a probe for the model's computational priorities. Our results highlight a gap between geometric theory and learned practice and identify a systemic \textit{Curvature Collapse} on the Long Range Graph Benchmark (LRGB). Rather than alleviating bottlenecks, the global attention mechanism exacerbates them, significantly increasing the prevalence of negative curvature (e.g., from 57\% to 84\% in \texttt{peptides-func}). Furthermore, our causal pruning experiments reveal a pathological reliance on these inefficient structures. The model creates a dependency on these congested corridors, effectively collapsing the global Transformer into a locally-biased MPNN that fails to leverage its theoretical capacity for long-range connectivity. Our contributions are as follows.
\begin{itemize}
    \item We perform a curvature-aligned MA analysis and use barbell graphs and \texttt{ZINC} / \texttt{Tox21} to test the bottleneck hypothesis, finding that attention extremes rarely over-represent negatively curved edges.
    \item We identify the \textit{Curvature Collapse} phenomenon, where global attention exacerbates rather than resolves topological bottlenecks.
    \item We demonstrate via causal pruning ablation studies that GTs develop a pathological reliance on these negatively curved regions, failing to capture long-range interactions, falling back into a locally-biased MPNN in the absence of explicit geometric guidance.
\end{itemize}

\section{Related Work}

\paragraph{Oversquashing and Curvature.} 
The limitations of message-passing GNNs in capturing long-range dependencies have been extensively studied, with oversquashing emerging as a failure mode where information from exponential receptive fields is compressed into fixed-size vectors \cite{alon2020bottleneck}. \citet{toppingunderstanding} proposed BFc as a metric for identifying topological bottlenecks, establishing a theoretical connection between bottlenecks and oversquashing. Recent position papers refine this link by disentangling \emph{topological} and \emph{computational} bottlenecks \cite{arnaiz2025oversmoothing} or arguing that oversquashing is often  irrelevant on standard benchmarks \cite{kormann2026position}. While curvature-based rewiring strategies have been proposed to alleviate these bottlenecks \cite{giraldo2023trade,toppingunderstanding}, recent studies have also questioned their consistent effectiveness across different tasks and hyperparameters \cite{torieffectiveness,tortorella2022leave}. Our work contributes to this debate by moving from proxy-driven interventions to model-internal evidence, analyzing where attention concentrates relative to bottlenecks and what the model functionally relies on.

\paragraph{Graph Rewiring Strategies.} 
To address the structural limitations of fixed input graphs, rewiring techniques explicitly modify connectivity to facilitate information propagation. Geometric approaches, such as Stochastic Discrete Ricci Flow (SDRF) \cite{toppingunderstanding}, operate by adding edges to negatively curved regions (bottlenecks) to \textit{flatten} the geometry of the graph. Spectral methods such as FOSR \cite{karhadkar2022fosr} optimize the spectral gap of the graph, utilizing the connection between Cheeger's inequality and mixing times to enhance global connectivity. Other strategies incorporate virtual nodes to create global shortcuts or leverage diffusion processes. Although these static interventions can theoretically alleviate oversquashing, they often rely on heuristics that may not align with the downstream task \cite{torieffectiveness,kormann2026position}. Furthermore, they decouple structural optimization from representation learning. This motivates the study of GTs, which can be viewed as learning a soft, dynamic rewiring (via the attention matrix) that is end-to-end differentiable and task-specific. A key open question, which we explore, is whether this learned rewiring recovers the geometric properties that rewiring methods aim to enforce.

\paragraph{Graph Transformers.}
To overcome the locality bias of MPNNs, Graph Transformers generalize the Transformer architecture to graphs by treating nodes as tokens and enabling all-to-all communication \cite{dwivedi2020generalization, kreuzer2021rethinking}. Approaches range from injecting structural encodings \cite{mialon2021graphit} to using spectral attention \cite{kreuzer2021rethinking}. Although theoretically capable of modeling global interactions, the actual connectivity patterns learned by these models and their interaction with the underlying graph topology remain underexplored. We specifically investigated whether these global mechanisms effectively \textit{flatten} the graph or fall prey to the same topological constraints as MPNNs.

\paragraph{Massive Activations.}
Recent interpretability research in Large Language Models (LLMs) has identified \enquote{Massive Activations}, extreme, heavy-tailed activation patterns, as consistent indicators of linguistic features and knowledge retrieval \cite{sun2024massive}. Extending this to the graph domain, \citet{bini2024massive} demonstrated that MAs in GNNs serve as reliable signals of domain-specific importance, such as functional motifs in molecules. In this work, we leverage MAs as a tool to decode the \textit{learned topology} of GTs, mapping these high-magnitude attention weights back to the graph's topology to understand the model's geometric preferences.

\section{Preliminaries}
\subsection{Graphs Neural Networks} In our analysis, we consider an undirected graph $G = (\mathbb{V},\mathbb{E})$. The graph is identified by a set $\mathbb{V}$ of nodes, which possess a feature vector $\mathbf{x}_i \in \mathbb{R}^{n_{0}}$, and a set of edges $\mathbb{E} \subset \mathbb{V} \times \mathbb{V}$. The adjacency matrix, which describes the connectivity of the graph, is denoted by $\textbf{A}$.

Given the graph $G$, we denote $\mathbf{h}^{(l)}_i$ as the representation of node $i$ at layer $l$, with $\mathbf{h}^{0}_{i} = \mathbf{x_i}$. Given differentiable layer-dependent functions $\phi^{l}$ and $\psi^{l}$, the message passing mechanism is defined as:
\begin{equation}\label{eq: message-passing}
    \mathbf{h}_i^{(l+1)}=\phi^{l}\bigg(\mathbf{h}_i^{(l)}, \sum_{j=1}^n \hat{\textbf{A}}_{i j} \psi^{l}\left(\mathbf{h}_i^{(l)}, \mathbf{h}_j^{(l)}\right)\bigg) \,,
\end{equation}
where $\hat{\textbf{A}}$ denotes a variation of the adjacency matrix (e.g., augmented with self-loops). 

\paragraph{Graph Transformer.}
We utilize standard Graph Transformer architectures (e.g., \cite{dwivedi2020generalization, mialon2021graphit, kreuzer2021rethinking}) that combine message passing with global attention mechanisms. Unlike MPNNs, which are restricted to local neighborhoods $\mathcal{N}(i)$, GTs allow node $i$ to attend to all other nodes $j \in \mathbb{V}$ in the graph. The update rule incorporates an attention-weighted aggregation of global features:
\begin{equation}\label{eq: attention}
    \mathbf{\hat{h}}_i^{(l+1)} = \sum_{j \in \mathbb{V}} \text{softmax}_j \left( \frac{\mathbf{Q} \mathbf{h}_i^{(l)} \cdot (\mathbf{K} \mathbf{h}_j^{(l)})^\top}{\sqrt{d_k}} \right) \mathbf{V} \mathbf{h}_j^{(l)}
\end{equation}
where $\mathbf{Q}, \mathbf{K}, \mathbf{V}$ are learnable query, key, and value projection matrices, and $d_k$ is the dimensionality of the key vectors. This global receptive field theoretically allows the model to capture long-range dependencies in a single layer, irrespective of the geodesic distance between the nodes. In many modern architectures (e.g., GraphiT \cite{mialon2021graphit}, GT \cite{kreuzer2021rethinking} and GraphGPS \cite{rampavsek2022recipe}), this global channel operates in parallel with a local MPNN channel to capture both local structure and global context.

\subsection{Graph Curvature}
Using discrete curvature notions to detect local bottlenecks in graphs stems from differential geometry, where  it is well known that the Ricci curvature describes whether two geodesics, which start close to each other, either diverge (negative curvature), stay parallel (zero curvature), or converge (positive converge). In graph topology, The graph analogs for these spaces are trees, four-cycles, and triangles. Discrete curvature notions, in essence, capture the occurrence of such structures around a given edge. Intuitively, negatively curved edges exhibit more tree-like structures in their local neighborhoods, which could lead to an exponentially growing receptive field and, therefore, a computational bottleneck.

We can compute \textit{Balanced Forman Curvature} (BFc) as follows. Let $G$ be an undirected graph from which we consider an edge $ i\sim j$ and denote by $d_i$, the degree of the node $i$, and $d_j$ the degree of the node $j$. The notions of discrete common curvature in graphs depend on the following topological aspects of the graph: (i) The common neighbors of node $i$ and node $j$ are denoted by $\sharp_{\triangle}(i,j)$, which correspond to the triangles located at the edge $i \sim j$. (ii) The neighbors of $i$ (resp. $j$) that form a four-cycle based on $i \sim j$ without diagonals inside are denoted by $\sharp_{\square}^{i}$ (resp. $\sharp_{\square}^{j}$) and (iii) the maximum number of four-cycles without diagonals inside that share a common node, which is denoted by $\gamma_{max}$. Discrete curvature notions combine these quantities to describe the local topology of the graph around the given edge $i \sim j$. BFc is then computed as \cite{toppingunderstanding}: \footnote{We denote by $x \vee y \doteq \mathrm{max}(x,\: y)$ (resp. $x \wedge y \doteq \mathrm{min}(x,\: y)$) the maximum (resp. minimum) of two real numbers.}
\begin{align}\label{eq: BFC_w4cycle}
        BFc(i,j) =& \frac{2}{d_i} + \frac{2}{d_j} - 2  + 2\frac{|\sharp_{\triangle}(i,j)|}{d_i \vee d_j} + \frac{|\sharp_{\triangle}(i,j)|}{d_i \wedge d_j} \\
                  &+ \frac{\left(|\sharp_{\square}^{i}| + |\sharp_{\square}^{j}|\right)}{\gamma_{max}( d_i \vee d_j)} \,. \nonumber
\end{align}
\subsection{Massive Activations Analysis}\label{sec: 3.3}
We adopt the definition of Massive Activations (MAs) proposed by \citet{bini2024massive}, which identifies them as statistically significant outliers in the weight distribution of the attention mechanism. These extreme activations are not merely numerical anomalies; in fact, MAs have been demonstrated to encode domain-relevant signals. For instance, in molecular graphs, MAs were found to predominantly localize on statistically common structural elements (e.g. single/double bonds) while consistently sparing rarer chemically informative features like triple bonds, effectively acting as \textit{attribution indicators} for informative edges.

Formally, we extract the attention weights $\alpha_{ij}^l$ for each edge $(i,j)$ in each layer $l$. Since attention distributions can vary in scale across layers and heads, we normalize them using the robust median statistic to compare the activation intensity across the model. We define the \textit{Activation Ratio} for an edge as:
\begin{equation}\label{eq: act_ratio}
    \text{ratio}(\alpha_{ij}^l) = \frac{|\alpha_{ij}^l|}{\text{median}_{k,m \in \mathbb{E}}(|\alpha_{km}^l|)}
\end{equation}
An edge is flagged as exhibiting massive activation if its maximum ratio across all heads and layers exceeds a high percentile threshold (typically the $95^{th}$ or $99^{th}$ percentile) of the aggregate distribution. This threshold allows us to isolate the \textit{heavy tail} of the attention mechanism; the specific subset of edges that the model preferentially prioritizes with extreme intensity. In this work, we leverage this framework to map these high-magnitude weights back to the graph's discrete curvature, effectively probing whether the model's learned structural bias aligns with topological theory.
\section{Experiments}
\subsection{Synthetic Graphs}\label{sec: barbell}
\begin{figure}[ht!]
    \centering
    \includegraphics[width=1\columnwidth]{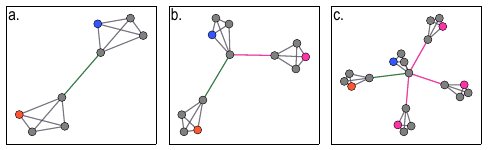}
    \caption{\textbf{The standard barbell graph (a.), the simple modified barbell graph (b.) and the extended modified barbell graph (c.) considered.} The source node (orange) contains a signal to be received by the target node (blue). The modified barbells add one or more dummy cliques containing nodes with no information about the task. These dummy cliques contain a dummy signal node (pink). The green bridge connects the source and target cliques (task-relevant), while pink bridges connect dummy cliques to the target (task-irrelevant).}
    \label{fig: barbell_toy_example}
\end{figure}

To illustrate the principle of our study, we analyze a simplified graph. We consider the barbell graph (with 4 nodes per clique) and its modified variants illustrated in Figure \ref{fig: barbell_toy_example}. In these graphs, we define the simple task of reconstructing a signal from the source node (orange) to the target node (blue). The modified variants of the barbell contain one (in the simple case) or three (in the extended case) additional cliques that connect with a bridge edge to the target node clique. Each of these dummy cliques also contain a source node, which is not the same signal as in the true signal node. The 16-dimensional source node feature is initialized at random, while the dummy signal nodes are initialized with a constant feature vector, unique for each dummy signal node. All other nodes have a zero vector as initial feature vector.

\begin{table}[ht]
\centering
\caption{Edge type encoding for the (modified) barbell graph.}
\label{tab:edge_types}
\begin{tabularx}{\columnwidth}{ccX}
\toprule
\textbf{Type} & \textbf{Name} & \textbf{Description} \\
\midrule
0 & $Cl$-$Cl$ & Between clique nodes \\
1 & $S$-$B_{S}$ & Source node to bridge node  \\
2 & $B_{T}$-$T$ & Bridge node to target node  \\
3 & $Bridge$ & Bridge edge (between source and target) \\
4 & $Bridge~Dummy$ & Dummy bridge edge (between dummy and target) \\
\bottomrule
\end{tabularx}
\end{table}
For each graph, we consider two different variations: topologically informative edge features and random edge features. In the case of topologically informative edge features, we attribute to each edge a scalar edge feature based on its location on the graph as described in Table \ref{tab:edge_types}. For random edge features, we randomly permute the edge features previously attributed to all edges. In this way, we maintain the distribution of features, but each edge is randomly assigned a feature. In both cases, the edge features are uninformative with respect to the task content as they do not contain information with regard to the message being passed. However, the topological correct features do discern between \enquote{useful} and \enquote{non-useful} edges for message-passing.

Using the formula for the Balanced Forman Curvature ($BFc$) we can compute that the bridge edge in the barbell graph has a negative curvature\footnote{For a barbell with cliques of size $k$, the bridge nodes have degree $k$ (connecting to $k-1$ nodes in the clique and a bridge neighbor). Since no triangle or $4$-cycles are formed, the $BFc$ reduces to $ \frac{4}{k} - 2$, which is negative for any non-trivial clique.}. These bridge edges are important from a geometic perspective, as a negative curvature is a measure for a topological bottleneck (making the important distinction that this does not necessarily imply a computational bottleneck/oversquashing \cite{arnaiz2025oversmoothing}). By analyzing the activation ratios developed during learning on these graphs, we analyze whether a learned attention mechanism also recognizes the relevance of these edges.

\paragraph{Experimental setup} We train a 3-layer local Graph Transformer to reconstruct the signal from the source node to the target node. As the two nodes are 3 hops away, a 3-layer model should propagate the information required to solve this task. For each version of the barbell graph, we generate $256$ graphs with node features as described above. The source and target nodes are identical for every graph, and the random edge features are permuted differently for each graph. Once trained, we extract the activations following the analysis in \cite{bini2024massive}, normalizing them based on the median per layer. The activation ratios are collected on a test dataset comprising 26 graphs (representing $10\%$ of the train set). The final train and MSE loss test are reported in Table~\ref{tab:barbell_results}, showing that the model has solved the task reasonably well (reaching a test of MSE $\approx 0.2$--$0.3$).
\begin{table}[ht!]
\centering
\caption{Signal reconstruction MSE on barbell graph variants (Train / Test).}
\label{tab:barbell_results}
\begin{tabular}{lcc}
\toprule
\diagbox[width=10.5em]{Graph}{Edge Features} & Topological & Permuted \\
\midrule
Standard          &  0.088 / 0.281   & 0.098 / 0.308  \\
Modified          &  0.106 / 0.286   & 0.097 / 0.323  \\
Extended Modified &  0.085 / 0.220   & 0.089 / 0.321 \\
\bottomrule
\end{tabular}
\end{table}

\begin{figure}[ht!]
    \centering
    \includegraphics[width=1\columnwidth]{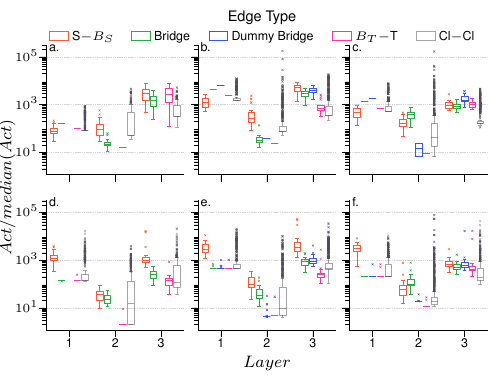}
    \caption{\textbf{Activation Ratios for the three barbell graphs: standard (a., d.), modified (b., e.) and extended modified (c., f.). The top row (a., b., c.) shows the ratios when the edge features are topologically accurate, while the bottom row (d., e., f.) when the features are permuted.} The dummy bridges in the modified and extended  activation ratios for the task-relevant bridge edge (green) and the dummy bridges (blue) in the modified datasets show the same (or higher) activation ratios. Note the log-scale on the $y$-axis. }%Distribution statistics are reported in Appendix \ref{Apdx: barbell}.}
    \label{fig: barbell}
\end{figure}

\paragraph{Results}
The resulting distributions for each type of edge (as classified in Table \ref{tab:edge_types}) are shown in Figure \ref{fig: barbell}  in all three barbell graph datasets and for both topologically accurate edge features, as well as randomly permuted ones. 

In layer 2, where the bridge edge plays a crucial role in the transport of information from the source clique to the target clique, we see that in three cases (Figure \ref{fig: barbell}, panel c., e, and f.) the activation ratios are significantly higher for the bridge edge relevant to the task than for the dummy bridges. In the case of the modified barbell with topological edge features (Figure \ref{fig: barbell} panel b.), both median values are similar, but the bridge edge ratios show larger variance. This low variance of the dummy bridges occurs for each of the four cases at layer 2. The Graph Transformer model has therefore learned which bottleneck matters, not that bottlenecks matter.
This variance asymmetry between task-relevant and -irrelevant bridges suggests that activations depend on the actual signal content, responding differently based on the signal being propagated. In contrast, dummy bridge activations remain stable across instances, consistent with the fact that they do not carry information relevant to the task. This pattern persists for both the topologically accurate and permuted edge features, indicating that the behavior is not driven by the edge feature distribution, where the sparser features are signaled regardless of the topology or task. 
Additionally, we note a second pattern from the edges of the internal cliques (Cl-Cl). Despite having no crucial role in information transfer between cliques, these edges show the highest outlier values for the activation ratio. The Cl-Cl edges are the most abundant in the graph and could play the role of attention sinks \cite{xiaoefficient}, where the model can deposit the attention mass. 

In barbell experiments, we find that even edges identified as topologically important by curvature are discerned based on their activation values at the critical information layer (i.e., layer 2). Additionally, since in this experiment we replicate 256 identical graphs, our findings reveal that activation variance distinguishes between task-relevant and -irrelevant edges that are topologically identical. However, the highest outliers in activation values (MAs) concentrate on abundant intra-clique edges at this critical layer, rather than task-relevant bridges. Through these controlled experiments, we establish that MAs do not necessarily track curvature-identified bottlenecks, but that topologically identical edges can be discerned by graph transformers based on their task relevance. In the next section, we extend our analysis beyond the topologically limited barbell graphs and study the activation behavior across diverse molecular datasets.

\subsection{Molecular Graph Datasets}\label{sec: molecular}
We extend our analysis to molecular graph datasets, namely \texttt{ZINC} \cite{irwin2012ZINC} and \texttt{Tox21} \cite{mayr2016deeptox,huang2016tox21challenge} on which we train Transformer architectures Graph Transformer \cite{dwivedi2020generalization}, GraphIT \cite{mialon2021graphit} and SAN \cite{kreuzer2021rethinking} with a local graph. The activation ratios are extracted on a test set after training. For \texttt{Tox21} the test dataset contains 7831 graphs, while for \texttt{ZINC} there are 1000. To save MA logs, we follow the original procedure as described in \cite{bini2024massive}.

On every molecular graph in the dataset we subsequently compute the \textit{Balanced Forman curvature} for each edge in the graph. These graph curvatures are then combined with the activation ratios. As highlighted in section \ref{sec: 3.3}, edges are flagged as a massive activation if their max ratio belongs to the $95^{th}$ percentile of ratios, instead of a fixed threshold. 

\begin{figure}[ht!]
    \centering
    \includegraphics[width=1\columnwidth]{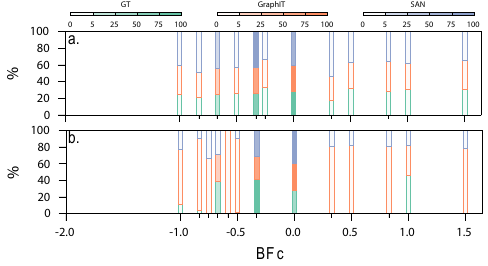}
    \caption{\textbf{Proportions of massively activated edges (based on 95th percentile) within each curvature value per model and per dataset ( (a.) \texttt{ZINC} (b.) \texttt{Tox21})} We show for each curvature value (\textit{x-axis}) the relative proportion (model-wise) of edges that have been designated as massively active. For each model we shade the region based on number of edges that are present relative to all massively activated edges of that model} %Distribution statistics are reported in Appendix \ref{Apdx: molecular}.}
    \label{fig: tox21_per_model}
\end{figure}

\paragraph{Distribution of MAs Across Curvature Values.} In Figure \ref{fig: tox21_per_model} we show the edges marked as massively activated in both datasets. The massively activated edges are binned on the basis of their curvature value. For each curvature bin we display a stacked color coded bar based on the model. The length of each segment in a bar denotes how many of the massively activated edges belong to the specific curvature value that was flagged by a model. For example, in the case of \texttt{Tox21} at $BFc = -1$ we see that GraphIT proportionally contains more masssively activated edges then SAN or GT. 

Additionally, for each of the bars we shade them based on the percentage they represent within all the massively activated edges when looking at edges per model (so an intensely colored bar means that most massively activated edges of that model are present at that curvature value). From the results, we can note that most massively activated edges are concentrated on a few curvature values. For both datasets, this occurs at $BFc = 0$, as well as at $BFc = -0.5$. This can be seen by the intensely colored bars at those curvature values. Interestingly, for those values, all models flagged around the same number of edges, as indicated by the segment lengths being similar for those bins.  For the \texttt{Tox21} dataset we see that the presence of massively activated edges at specific curvature values is model dependent, with GraphIT dominating certain curvature regions.

\paragraph{Enrichment Analysis}
To quantify whether MAs over or under-represent specific curvature regions relative to the base distribution, we computed the enrichment ratio (Eq.~\eqref{eq: Enrichment}). As both the \texttt{Tox21} and \texttt{ZINC} datasets have discrete curvature values, we do not need to bin the $BFc$ values before computing the enrichment ratios. Given a curvature value $\tilde{c}$ we define the enrichment as
\begin{align}\label{eq: Enrichment}
E(\tilde{c})=\frac{P\!\left(BFc(e)=\tilde{c} \mid e\in \mathbb{E}~\And~e \in MA\right)}{P\!\left(BFc(e)=\tilde{c}\right)} \,.
\end{align}
Values above 1 for the enrichment indicate an over-representation of MAs at that curvature value based on what one would expect when considering the underlying curvature distribution. 
\begin{figure}[ht!]
    \centering
    \includegraphics[width=1\columnwidth]{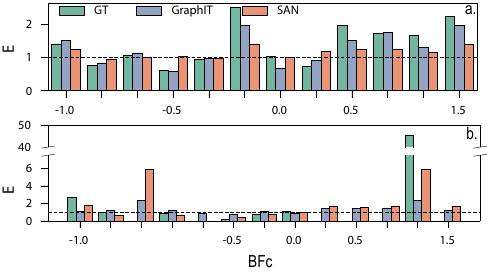}
    \caption{\textbf{Enrichment values for all three models over both datasets: \texttt{ZINC} (a.) and \texttt{Tox21} (b.).} The enrichment values show that the relation between curvature and massive activations is non-monotonic, where topological bottlenecks (indicated by extreme negative curvatures) are marginally over-represented. Additionally, positively curved edges are also over-represented. The dotted line indicates $E = 1$ meaning no over- or underrepresentation.}
    \label{fig: enrichment}
\end{figure}

\begin{figure}[ht!]
    \centering
    \includegraphics[width=1\columnwidth]{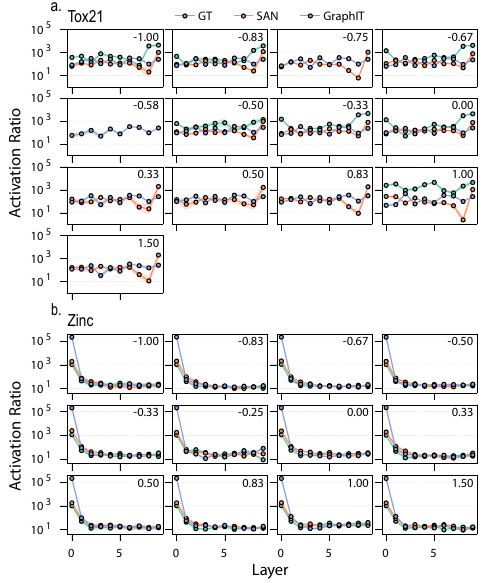}
    \caption{\textbf{Layer-wise evolution of activation ratios of MAs across curvature bins for \texttt{Tox21} (a) and \texttt{ZINC} (b).} Each panel corresponds to a BFc value and lines show mean activation ratio per layer. \texttt{ZINC} exhibits uniform early-layer decay across all curvature values, while \texttt{Tox21} maintains elevated ratios throughout, with GT showing late-layer spikes at negative curvatures.} 
    \label{fig: layer_ma_evolution}
\end{figure}

The enrichment values, presented in Figure \ref{fig: enrichment}, highlight the non-monotonic relation between massive activations and curvature. Most importantly, we find that massively activated edges predominantly do not aggregate only on topological bottlenecks (as indicated by negative curvature). The model therefore does not disproportionately attend geometric bottlenecks relative to their frequency. For \texttt{ZINC} (Figure \ref{fig: enrichment}, panel a.), the enrichment of extreme curvature values is small for all three transformer models, with less extreme negative curvatures even being under-represented. For \texttt{Tox21} (Figure \ref{fig: enrichment}, panel b.) we see similar patterns, which are heavy outliers at both positive ($BFc = 1$) and negative ($BFc = -0.75$) curvature. The heterogeneous enrichment pattern, spanning both bottleneck and non-bottleneck regions, further demonstrates that MA placement is driven by factors beyond topological constraints, likely reflecting task-specific chemical features.
\paragraph{MA evolution over layers}
Finally, in analogy to the per-layer analysis of the barbell graph, we can look at how the activation ratio evolves for MAs in molecular datasets. Figure \ref{fig: layer_ma_evolution} shows how the activation ratios of the MA-labeled edges evolve between layers for the edges grouped by their BFc value. For \texttt{ZINC} (Figure \ref{fig: layer_ma_evolution}, panel b.) in all curvature bins, the activation ratios show a sharp decay from layer 0 to layer 1-2. The dominance of the first layer suggests that the importance of the edge is established in the first layer and subsequently normalized, with no curvature-specific modulation in deeper layers. In contrast, \texttt{Tox21}  (Figure \ref{fig: layer_ma_evolution}, panel a) shows activation ratios that remain elevated ($\sim10^{2}-10^{3}$) throughout all layers, with notable variation across the curvature bins. 

Through sections \ref{sec: barbell} and \ref{sec: molecular} we establish that attention mechanisms in Graph Transformers do not inherently align with geometric bottleneck theory. In a controlled experiment on the barbell graphs, we show that attention can distinguish task-relevant from task-irrelevant edges even when they share identical curvature. Extending to real molecular graphs, we find no consistent enrichment of MAs on topological bottlenecks. MA placement appears unrelated to the topology of the edge, indicating that Graph Transformers do not preferentially highlight bottlenecks. However, from a theoretical perspective, the link between negative curvature and information flow constraints remains valid. We now ask: if attention is geometry-blind, what happens on tasks where long-range dependencies are essential?
\subsection{LRGB Results}
To answer this question we extend our analysis to the Long Range Graph Benchmark (LRGB) dataset \cite{dwivedi2022long}, specifically \texttt{peptides-func} and \texttt{peptides-struct}, specifically designed to test the ability of GNNs to capture long range dependencies. Our geometric surgery reveals two distinct pathological behaviors composed of the attention mechanism, confirming the hypothesis of an \enquote{Attention Curse}.

\begin{figure}[ht!]
    \centering
    \includegraphics[width=0.92\columnwidth]{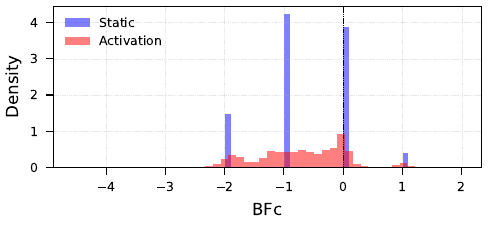}
    \caption{\textbf{Shift in Effective Geometry.} The distribution of Activation-Weighted Curvature (Red) is shifted significantly to the left compared to the Static Curvature (Blue), indicating a system-wide increase in negative curvature (bottlenecks).}
    \label{fig:curvature_distribution}
\end{figure}

\paragraph{The Curvature Collapse.} In both datasets, the Transformer mechanism fails to \enquote{flatten} the graph geometry as intended. Instead, it exacerbates the bottlenecking problem.  \cref{fig:curvature_distribution} and \cref{tab:curvature_collapse} compare the $BFc$ distribution of the original graph (with a predefined adjacency matrix) and graph with reweighted adjacency matrix based on the activation scores developed during training (see Appendix \ref{Apdx: lrgb_analysis} for details). 
\begin{table}[ht!]
%\small
\footnotesize
\centering
\caption{Prevalence of Negative Curvature Edges. The significant increase showcases the "Curvature Collapse" phenomenon.}
\label{tab:curvature_collapse}
\begin{tabular}{lcc}
\toprule
\textbf{Dataset} & \textbf{Static Graph} & \textbf{Activation Graph} \\
\midrule
\texttt{peptides-func} & 57\% & 84\% \\
\texttt{peptides-struct} & 57\% & 82\% \\
\bottomrule
\end{tabular}
\end{table}

This \textit{curvature collapse} indicates that the attention mechanism is concentrating the information flow into narrow, highly curved corridors rather than creating global shortcuts. The \enquote{maze} of the graph effectively becomes harder to navigate after the application of attention.
To test the reliance of the model on this new graph we perform a controlled pruning experiment. We categorized the edges into three sets (based on MA and curvature) and measure the increase in the validation loss when each set was physically removed from the graph by setting its corresponding adjacency matrix entry to zero. As shown in \cref{tab:causal_pruning}, both tasks exhibit similar pathological dependence on these bottleneck structures. 
\begin{table}[ht]
\centering
\caption{Impact of Causal Pruning on Loss. Retrieving Set A (MA on bottlenecks) consistently hurts performance, confirming the model relies on these inefficient structures. Details in \cref{Apdx: lrgb_analysis}.}
\label{tab:causal_pruning}
\resizebox{\columnwidth}{!}{%
\begin{tabular}{lccc}
\toprule
\textbf{Pruning Target} & \textbf{Region Type} & \textbf{Func ($\Delta$ Loss)} & \textbf{Struct ($\Delta$ Loss)} \\
\midrule
Set A (MA + Neg) & Bottleneck & \textcolor{red}{+0.1124} & \textcolor{red}{+0.1327} \\
Set B (MA + Pos) & Grid/Flat & +0.0153 & +0.0302 \\
Set C (NoMA + Neg) & Bottleneck & +0.0506 & +0.0790 \\
\bottomrule
\end{tabular}
}
\end{table}

\textit{1. Peptides-Func (Inefficient Dependency):} Removing MAs located in bottleneck regions causes a significant spike in loss ($+0.1124$, $\sim22\%$ relative error). This indicates that the model \textit{relies} on these connections. However, the increased negative curvature suggests that this reliance is geometrically inefficient and leads to force traffic through congested bottlenecks rather than building bypasses.

\textit{2. Peptides-Struct (Structural Overfitting):} We similarly find a massive loss spike ($+0.1327$, $\sim27\%$ relative error) when removing bottleneck MAs. This confirms that the \enquote{Attention Curse} is not a localized phenomenon but a consistent failure mode: the model consistently identifies and reinforces the graph's topological bottlenecks instead of resolving them.

These results suggest that without explicit geometric regularization, global attention mechanisms consistently degrade the graph topology, leading to inefficient routing and reliance on structural bottlenecks.

\paragraph{Global Connectivity Failure.} A consequence of this curvature collapse is the degradation of long-range interaction capabilities. In both datasets, the emergence of negative curvature bottlenecks implies that the global Transformer has functionally collapsed into a locally-biased MPNN. Instead of bridging distant nodes through flat or positively curved edges, the attention mechanism reinforces local highly curved structures. This effectively neutralizes the theoretical advantage of the global attention layer, rendering it unable to bypass the structural limitations of the underlying graph and forcing it to operate within a congested, short-range regime. The global Transformer effectively ignores the long-range interactions it was designed to capture. This effect could also explain why GTs fail to outperform standard GNNs on the LRGB dataset \cite{tonshoff2024where}.

\section{Discussion}
Our investigation into the interplay between Graph Transformer attention mechanisms and graph topology reveals a complex and often counter-intuitive relationship. Contrary to the theoretical expectation that global attention should alleviate topological bottlenecks by creating shortcuts, our findings suggest a more nuanced reality where learned importance frequently decouples from geometric significance.

\paragraph{The Disconnect Between Curvature and Importance.} 
Our experiments on synthetic barbell graphs demonstrate that while Graph Transformers can solve tasks requiring long-range information propagation, they do not necessarily prioritize topologically critical edges (bridges) in the way geometric theory would predict. We observed that Massive Activations often concentrate on abundant intra-clique edges rather than the bottlenecks themselves. This suggests that the model's learned topology is driven more by task-specific signal relevance than by intrinsic graph geometry. The fact that indistinguishable dummy bridges could be filtered out on the basis of signal variance further underscores that attention mechanisms are highly sensitive to the content of the message, not just the conduit.

\paragraph{Curvature Collapse as a Feature of Learning.}Perhaps most strikingly, our analysis of the LRGB benchmarks uncovers the phenomenon of \textit{Curvature Collapse}. Rather than \enquote{flattening} the graph to improve reachability, the attention mechanism systematically exacerbates negative curvature, effectively making the learned latent graph \textit{more} bottlenecked than the input structure. This paradox, i.e. a mechanism designed for global connectivity which ends up reinforcing local congestion, challenges standard assumptions about how GTs operate. Our causal pruning ablation studies confirm that this is not a benign artifact but a functional dependency since models actively rely on these congested corridors to perform.

\paragraph{The Functional Downgrade to MPNN.}
Our causal pruning ablation studies provide definitive evidence that this curvature collapse is not merely structural but functional. When MAs associated with negative curvature are removed, performance drops significantly, indicating that the model's computation is largely confined to these bottleneck regions. This implies a systemic \textit{downgrade} of the GT architecture: instead of transcending the graph's geometry via global attention, the presence of MAs induces a negatively curved manifold that restricts the model to a local message-passing regime. The Transformer, in effect, learns to behave like a standard MPNN, failing to realize its potential for solving long-range dependencies.

\paragraph{Implications for Model Design.}
The observation that MAs act as natural \textit{attribution indicators} for common substructures (as seen in molecular graphs) provides a dual perspective. On the one hand, it validates MAs as meaningful signals rather than noise. On the other hand, it highlights a potential failure mode where models over-fit to statistically dominant but chemically less informative features (like single bonds) at the expense of rarer, critical structures. This raises questions about whether explicit geometric guidance could help. Relying solely on the attention mechanism to discover optimal routing paths appears insufficient; without guidance, the optimization landscape seems to bias towards reinforcing existing topological constraints. 

\section{Future Works}
Our findings open several avenues for future research in Graph Transformer design and GNN interpretability. The disconnect between curvature-identified bottlenecks and MA placement suggests that topology analysis may serve better as a diagnostic tool than a prescriptive guide. The barbell experiments hint that learned attention encodes information beyond pure topology, but a systematic characterization of this relationship remains open.
 
However, the curvature collapse phenomenon suggests that some form of geometric guidance may still be necessary. Future architectures could incorporate curvature-aware loss functions that penalize the formation of negatively curved bottlenecks in the latent space, encouraging the model to maintain a \enquote{flatter} geometry that genuinely facilitates long-range communication. Moreover, the connection between MAs and functional downgrades warrants a broader analysis. It remains to be seen whether this behavior is universal across different attention mechanisms (e.g., linear or sparse attention) or specific to the softmax-based formulation. 

Finally, the role of MAs as intrinsic attribution indicators offers a promising direction for explainable AI. The observation that MAs concentrate on abundant structural elements while sparing rarer informative features provides a gradient-free probe into model behavior. Leveraging the localization of MAs could enable researchers to identify not just \textit{what} a model predicts, but \textit{where} in the graph it allocates its computation.
\section*{Impact Statement}
This paper presents work whose goal is to advance the field of Machine
Learning. There are many potential societal consequences of our work, none
of which we feel must be specifically highlighted here.
%\newpage
\bibliography{main_bib}
\bibliographystyle{icml2026}

%%%%%%%%%%%%%%%%%%%%%%%%%%%%%%%%%%%%%%%%%%%%%%%%%%%%%%%%%%%%%%%%%%%%%%%%%%%%%%%
%%%%%%%%%%%%%%%%%%%%%%%%%%%%%%%%%%%%%%%%%%%%%%%%%%%%%%%%%%%%%%%%%%%%%%%%%%%%%%%
% APPENDIX
%%%%%%%%%%%%%%%%%%%%%%%%%%%%%%%%%%%%%%%%%%%%%%%%%%%%%%%%%%%%%%%%%%%%%%%%%%%%%%%
%%%%%%%%%%%%%%%%%%%%%%%%%%%%%%%%%%%%%%%%%%%%%%%%%%%%%%%%%%%%%%%%%%%%%%%%%%%%%%%
\newpage
\appendix
\onecolumn
\section{Barbell Experiment}\label{Apdx: barbell}
To isolate the behavior of attention mechanisms on bottlenecks, we designed a synthetic Barbell graph task.

\paragraph{Graph Construction.}
The basic barbell graph consists of two fully connected cliques of size $N=4$, connected by a single "bridge" edge. To introduce distraction, we generate variants with additional "dummy" cliques also connected to the target clique via bridges.
\begin{itemize}
    \item \textbf{Basic Barbell:} Two cliques (Source and Target) connected by a bridge.
     \item \textbf{Extended Barbell:} One Source clique, one Target clique, and multiple Dummy cliques (1 or 3), each connecting to the Target clique via its own bridge.
\end{itemize}

\paragraph{Task Definition.}
The task is a signal reconstruction problem.
\begin{itemize}
    \item \textbf{Input:} Each node has a 16-dimensional feature vector.
    \item \textbf{Source Node:} One node in the Source clique is initialized with a random 16-dim signal vector.
    \item \textbf{Dummy Signal Nodes:} A node in each Dummy clique is initialized with a distinct, constant signal vector.
    \item \textbf{Target Node:} One node in the Target clique (the one connected to the bridge) serves as the target.
    \item \textbf{Other Nodes:} Initialized with zero vectors.
    \item \textbf{Goal:} The model must learn to transport the signal from the Source node to the Target node.
\end{itemize}

\paragraph{Edge Features.}
Edges are encoded with discrete types to potentially guide the model (or confuse it in the random case).
\begin{table}[h]
\centering
\caption{Edge type encoding for the Barbell graph.}
\label{tab:edge_types_appendix}
\begin{tabular}{ccl}
\toprule
\textbf{Type} & \textbf{Name} & \textbf{Description} \\
\midrule
0 & $Cl$-$Cl$ & Intra-clique edge \\
1 & $S$-$B_{S}$ & Source node to bridge-connected node \\
2 & $B_{S}$-$T$ & Bridge-connected node to Target node \\
3 & $Bridge$ & The critical bottleneck edge \\
4 & $Bridge~Dummy$ & Bridge to a dummy clique \\
\bottomrule
\end{tabular}
\end{table}
This setup tests the model's ability to distinguish the true bottleneck (Type 3) from dummy bottlenecks (Type 4) and intra-clique edges, based purely on the signal content or the provided structural features.
%\section{Molecular Graphs}\label{Apdx: molecular}
\section{Datasets}
We evaluated our hypothesis on the Long Range Graph Benchmark (LRGB) \cite{dwivedi2022long}, specifically focusing on the Peptide datasets which are designed to test the ability of GNNs to model long range interactions.

\paragraph{Peptides-Functional (\texttt{peptides-func}).}
This dataset consists of 15,535 peptides derived from the SATPdb database. The task is a multi-label graph classification problem with 10 classes labeled by peptide function (e.g., antifungal, anticancer, antibiotic).
\begin{itemize}
    \item \textbf{Graph Size:} Average of 150.94 nodes and 307.30 edges per graph.
    \item \textbf{Long-range property:} Functions are often determined by the peptide's global structure and motifs that may be distant in the 2D bond graph but close in the 3D space, requiring the model to capture long-range dependencies.
\end{itemize}

\paragraph{Peptides-Structural (\texttt{peptides-struct}).}
Sharing the same set of 15,535 peptide graphs as \texttt{peptides-func}, this dataset poses a graph regression task. The goal is to predict 11 structural properties computed from the molecule's 3D configuration (e.g., Inertia Mass, Spherocity, Plane of Best Fit).
\begin{itemize}
    \item \textbf{Challenge:} Since the input is only the 2D molecular graph (nodes=atoms, edges=bonds), the model must infer 3D structural properties solely from 2D topology, a task that inherently requires aggregating information across the entire graph.
\end{itemize}
Both datasets are divided into training sets (70\%), Validation (15\%), and tests (15\%), following the same procedure as \citet{dwivedi2022long}, while using scaffold splitting to ensure generalization to novel chemical structures.

\paragraph{\texttt{ZINC}.}
We use the standard subset of the \texttt{ZINC} dataset \cite{irwin2012ZINC, hu2020open}, consisting of 12,000 molecular graphs (10k train, 1k val, 1k test). The task is to perform graph regression to predict the constrained solubility (logP) of the molecule.
\begin{itemize}
    \item \textbf{Graph Size:} Small graphs, averaging ~23 nodes and ~50 edges.
    \item \textbf{Relevance:} A widely used benchmark for testing GNN expressivity and their ability to model basic chemical features.
\end{itemize}

\paragraph{\texttt{Tox21}.}
The \texttt{Tox21} dataset from the Open Graph Benchmark (OGB) \cite{huang2016tox21challenge} contains 7,831 molecular graphs. The task is multi-task binary classification across 12 toxicity labels.
\begin{itemize}
    \item \textbf{Graph Size:} Average of 18.5 nodes and 19.3 edges per graph.
    \item \textbf{Relevance:} Represents a real-world molecular property prediction task where identifying specific functional groups (toxicophores) is crucial.
\end{itemize}

\section{Details on LRGB Analysis}\label{Apdx: lrgb_analysis}
In this section, we provide a comprehensive breakdown of our experimental findings on the \texttt{peptides-func} dataset. Our results uncover a striking phenomenon termed the \textit{Attention Curse}: contrary to the expectation that global attention should \enquote{flatten} the graph geometry to facilitate long-range signal propagation, the learned attention mechanism actively warps the geometry into a more negatively curved, bottlenecked state.
\begin{figure}
    \centering
    \includegraphics[width=0.8\linewidth]{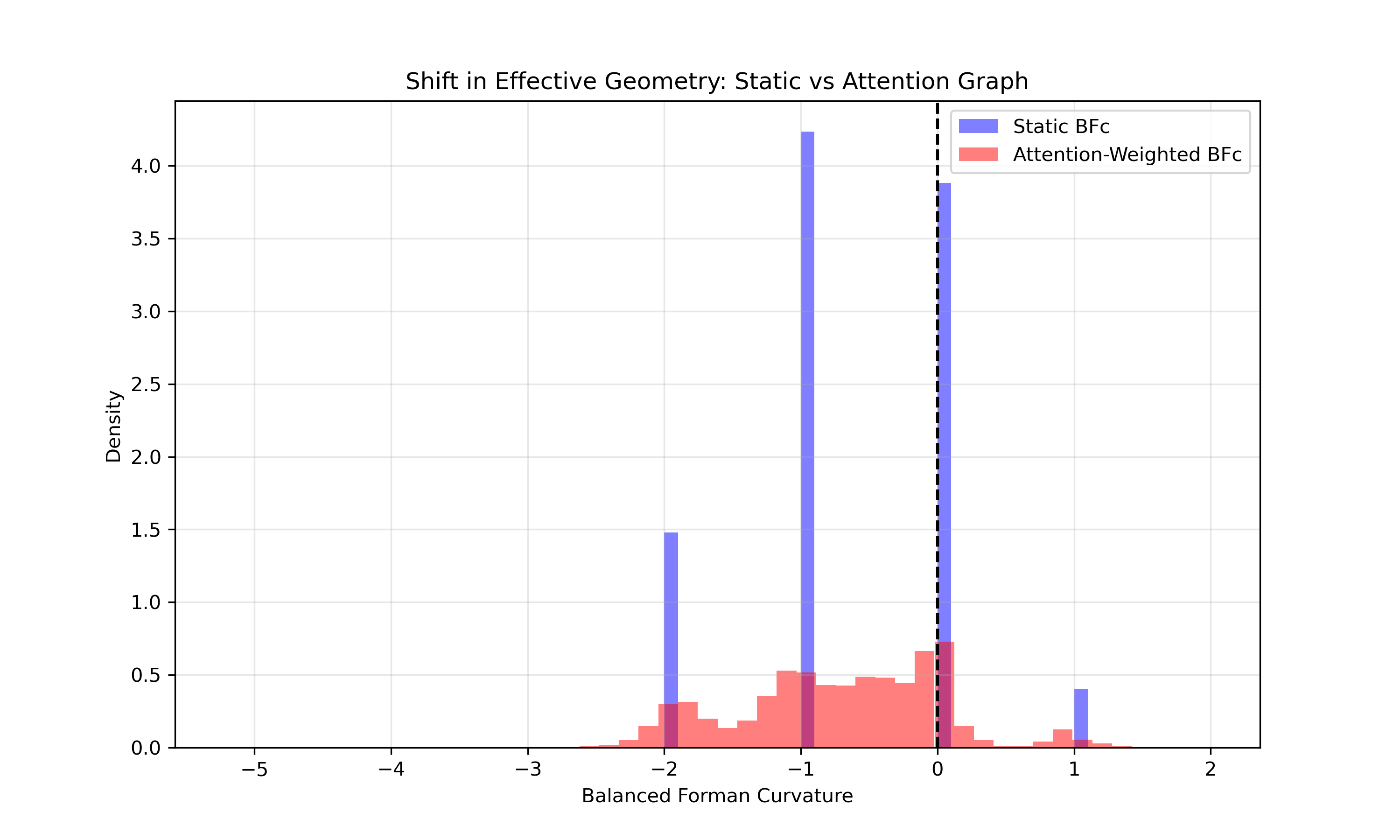}
    \caption{\textbf{The Geometry Shift.} The distribution of edge curvature weights shifts significantly to the left (more negative) in the Activation Graph (Red) compared to the Static Graph (Blue). This visualizes the \textit{Curvature Collapse}.}
    \label{fig:curvature_dist}
\end{figure}
\begin{figure}
    \centering
    \includegraphics[width=0.4\linewidth]{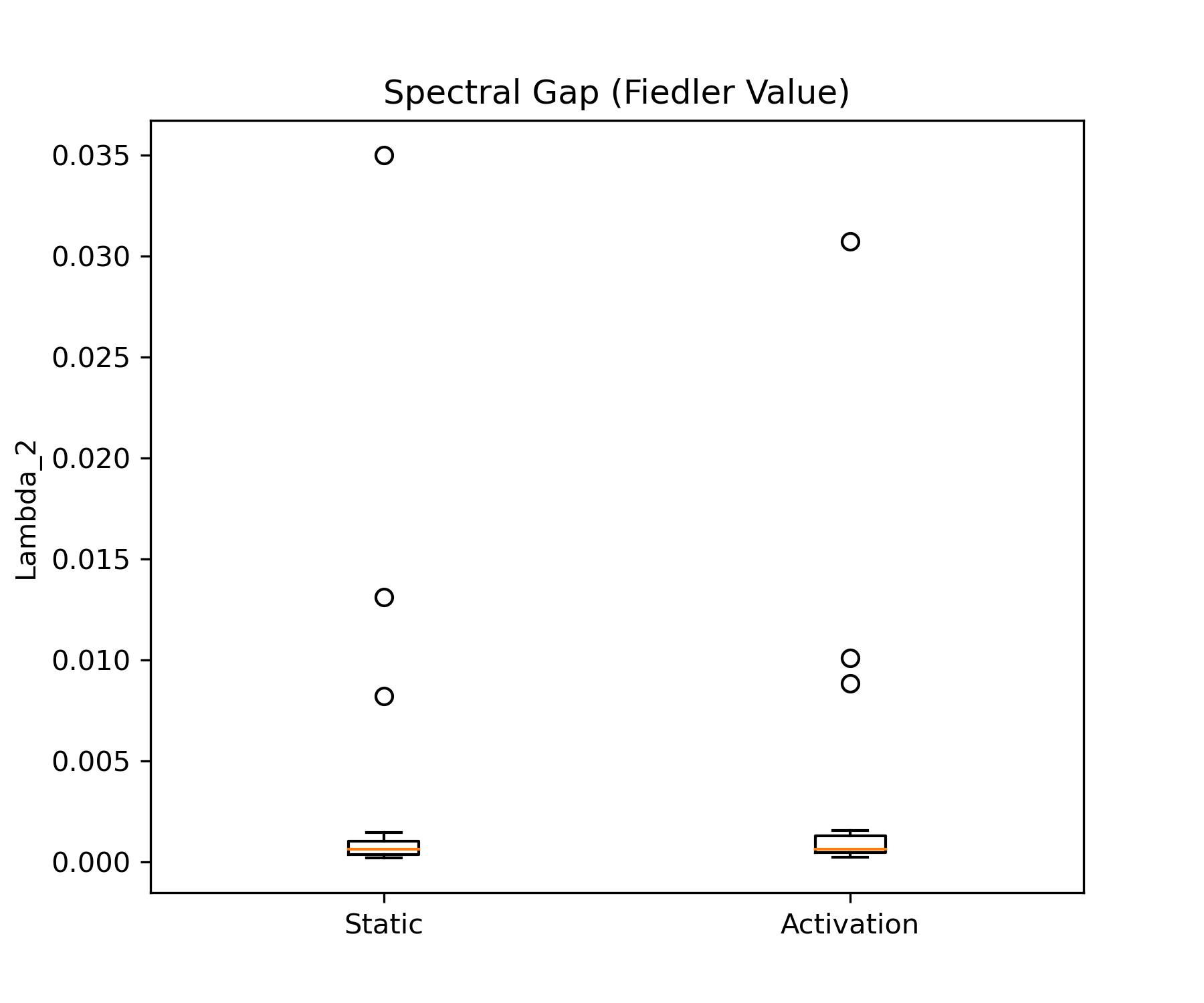}
    \caption{\textbf{Spectral Gap Comparison.} The Activation Graph (Red) consistently exhibits a lower spectral gap than the Static Graph (Blue), indicating reduced global connectivity.}
    \label{fig:spectral_gap}
\end{figure}
\begin{figure}
    \centering
    \includegraphics[width=0.4\linewidth]{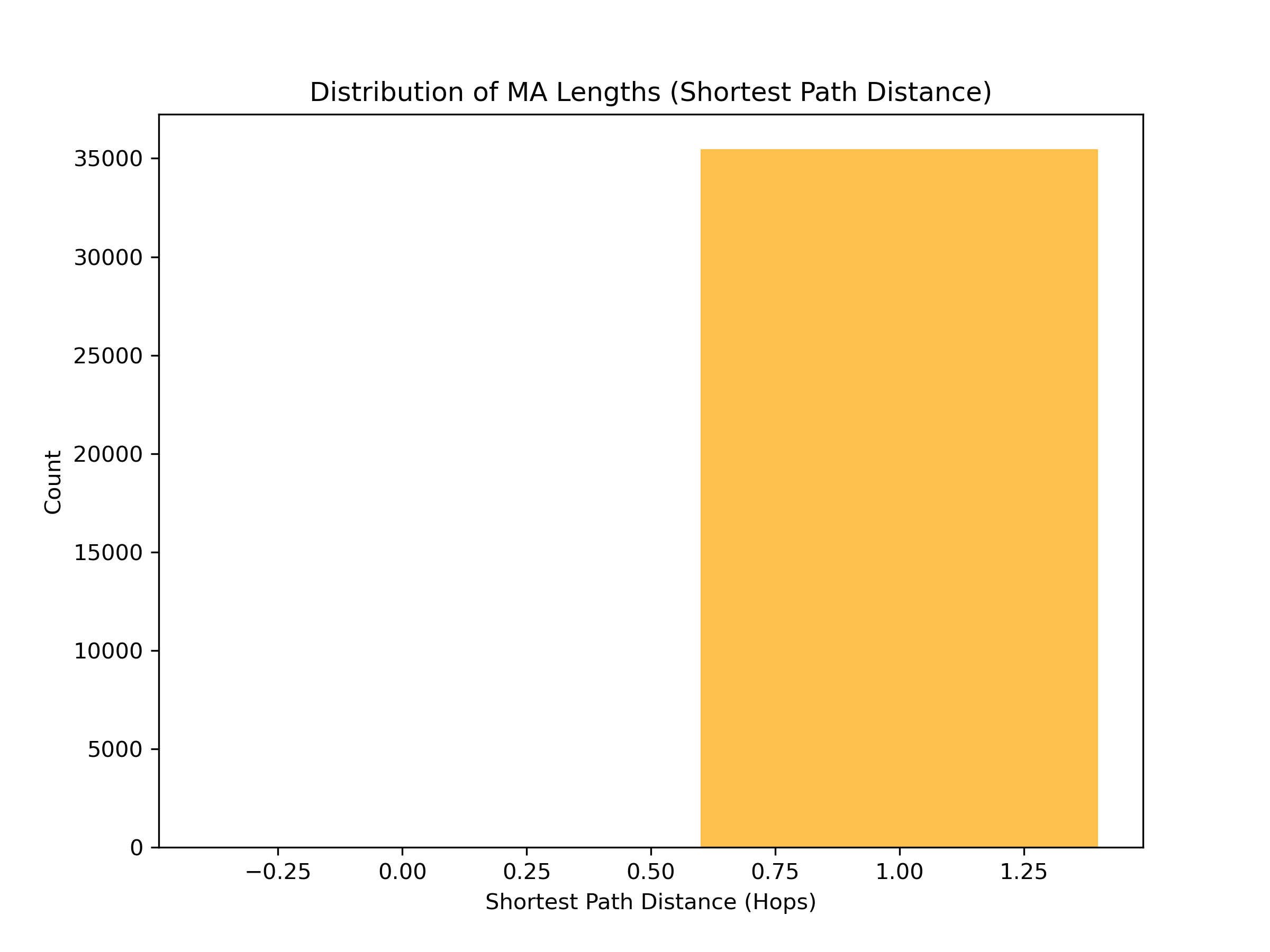}
    \caption{\textbf{MA Length Distribution.} MAs are overwhelmingly concentrated on short-range (0-1 hop) interactions.}
    \label{fig:ma_lengths}
\end{figure}
\subsection{The Curvature Collapse}
We quantified the topological impact of the learned attention mechanism using Balanced Forman Curvature (BFc).
\begin{itemize}
    \item \textbf{Static Graph Metric:} The baseline input graphs have a Weighted BFc of -0.6784, with 57\% of edges exhibiting negative curvature. This indicates a naturally bottlenecked structure.
    \item \textbf{Activation Graph Metric:} After training, the effective geometry induced by the attention mechanism (Massive Activations) drops to a Weighted BFc of -0.7008, with a staggering \textbf{84\%} of edges becoming negatively curved.
\end{itemize}
\textbf{Implication:} The model has increased the prevalence of negative curvature by roughly 27 percentage points. Instead of alleviating bottlenecks, it is amplifying them, concentrating flow into narrow, highly congested corridors.

\subsection{Causal Pruning Analysis}
To determine whether this curvature shift was functional or merely an artifact, we performed causal pruning ablation studies. We categorized the edges into three sets and measured the increase in the validation loss when each set was physically removed from the graph  by setting its corresponding adjacency matrix entry to zero (pruned).
\begin{enumerate}
    \item \textbf{Set A: MA + Negative Curvature (+11.2\% Relative Error).}
    These are Massive Activations occurring in bottleneck regions. Removing them causes the largest spike in loss (0.1124 increase over ~0.51 baseline). This confirms that the model is actively trying to route information through these bottlenecks but does so in a way that structurally degrades the graph.
    
    \item \textbf{Set B: MA + Positive Curvature (+1.5\% Relative Error).}
    MAs in "easy", grid-like regions. Removing them has minimal impact (+0.0153). This suggests substantial inefficiency: the model squanders its global attention budget on local, easy-to-reach neighborhoods where standard message passing would suffice.
    
    \item \textbf{Set C: NoMA + Negative Curvature (+5.1\% Relative Error).}
    Non-MA structural edges in bottlenecks. The significant impact of removing these "silent" edges (+0.0506) reveals that despite the global attention, the model remains critically dependent on the underlying local topology. It has not learned to bypass the bottlenecks; it is merely adding noise on top of them.
\end{enumerate}

\subsection{Visual Diagnostics}
We support our quantitative findings with visual evidence of structural degradation.
\begin{figure}
    \centering
    \begin{subfigure}[b]{0.48\textwidth}
        \includegraphics[width=\textwidth]{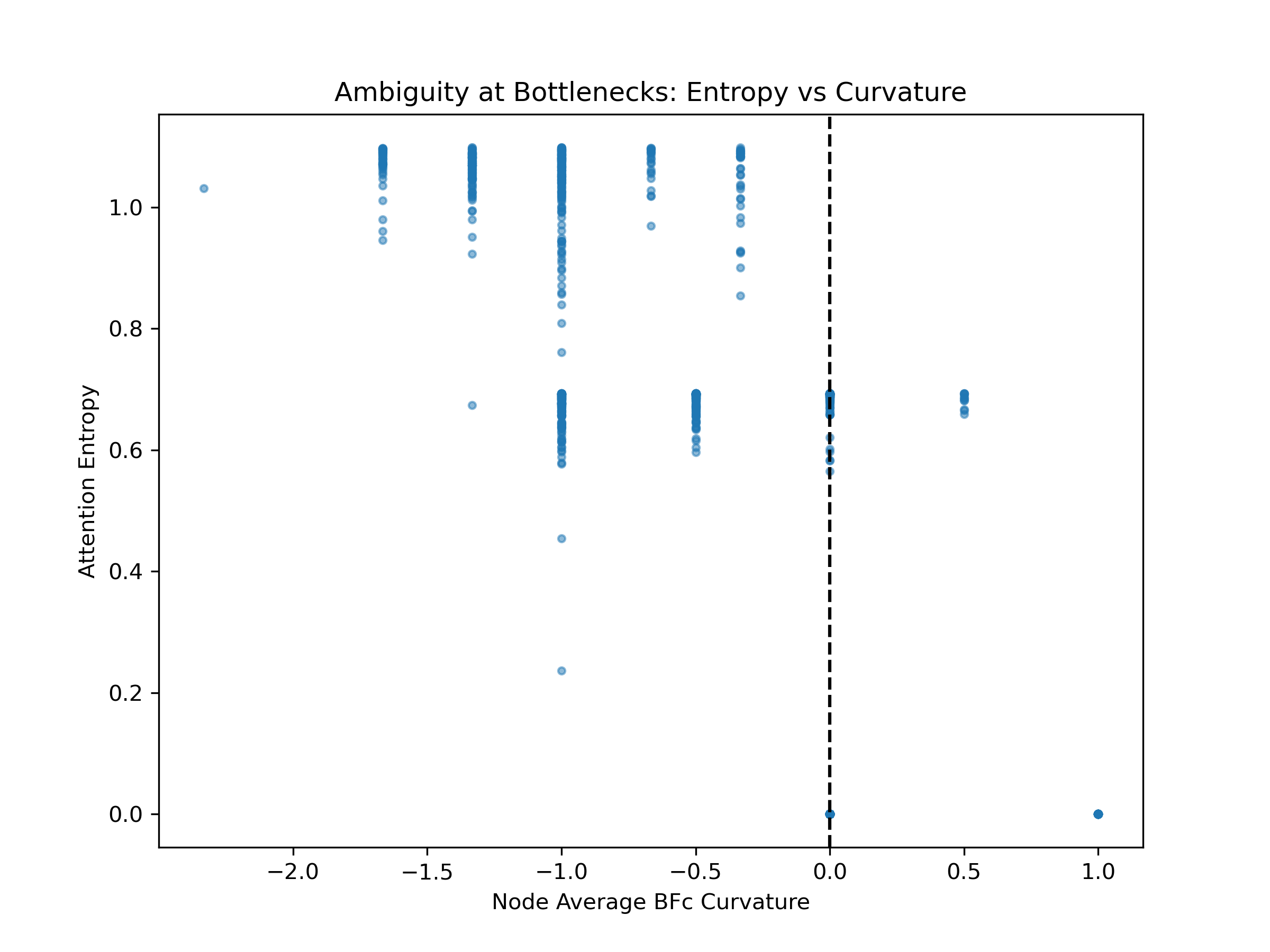}
        \caption{Entropy vs Curvature}
        \label{fig:entropy_curvature}
    \end{subfigure}
    \hfill
    \begin{subfigure}[b]{0.48\textwidth}
        \includegraphics[width=\textwidth]{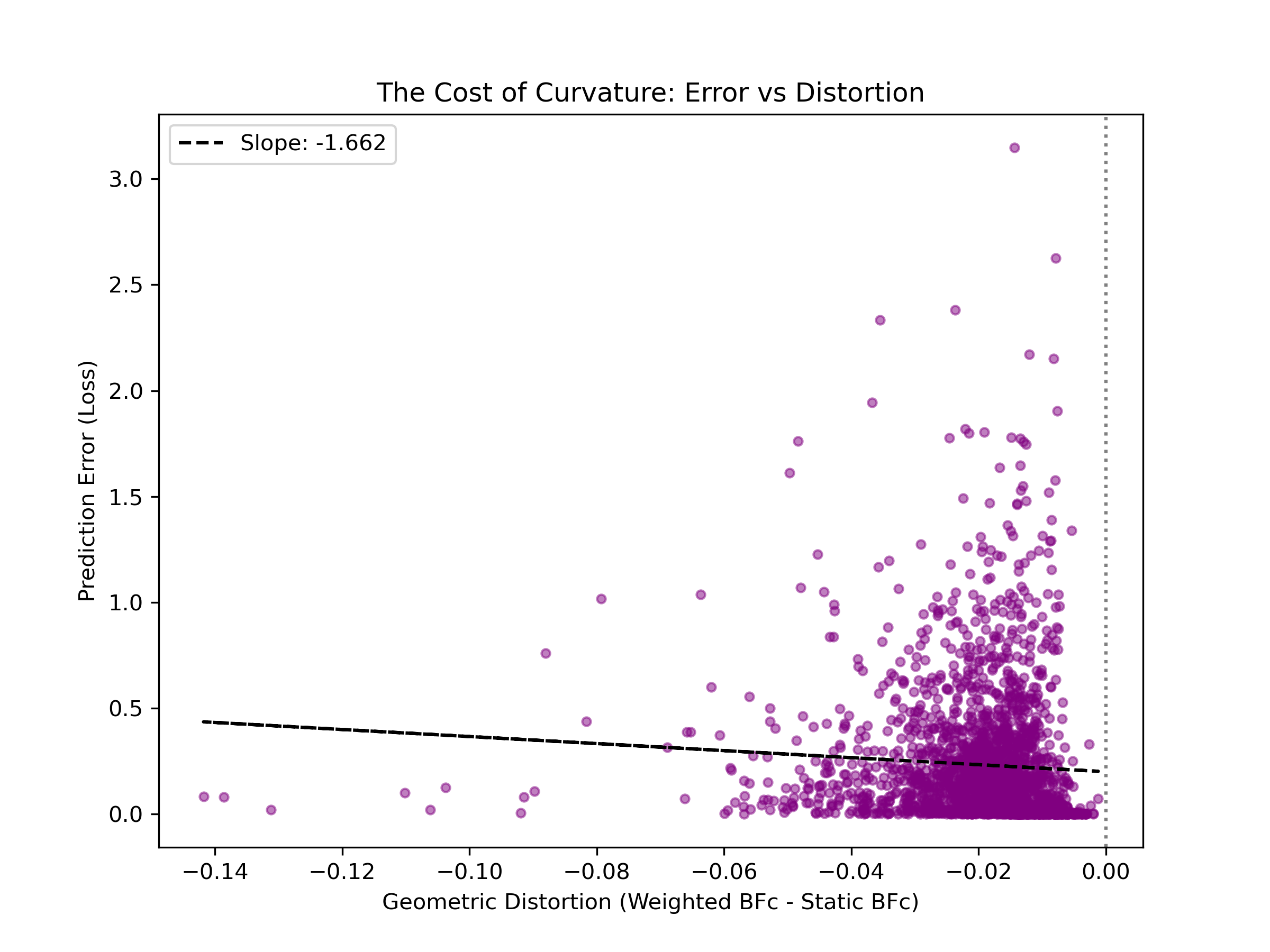}
        \caption{Error vs Geometric Distortion}
        \label{fig:error_distortion}
    \end{subfigure}
    \caption{\textbf{Diagnostic Plots.} (a) High entropy persists at negative curvature. (b) Geometric distortion correlates with higher error (Slope: -1.662).}
\end{figure}

\paragraph{Global Connectivity Failure.}
As shown in Figure \ref{fig:spectral_gap}, the spectral gap ($\lambda_2$) of the Activation Graph is consistently lower than that of the Static Graph. A lower spectral gap implies that the graph is easier to cut into disconnected components, indicating worse global connectivity. The global attention is functionally disconnecting the graph.

\paragraph{Local vs. Global Focus.}
Figure \ref{fig:ma_lengths} reveals that the vast majority of MAs span only 0 or 1 hops. There is almost no long-range attention mass. The Transformer has effectively collapsed into a locally-biased MPNN, ignoring the long-range interactions it was designed to capture.

\paragraph{Confusion and Cost.}
Finally, we observe high entropy in negatively curved regions (Figure \ref{fig:entropy_curvature}), suggesting that the model is confused in bottlenecks. Crucially, this geometric degradation correlates directly with performance: graphs with higher geometric distortion (more negative curvature shift) tend to have higher prediction errors (Figure \ref{fig:error_distortion}).

\end{document}